\pdfoutput=1
\documentclass[runningheads]{llncs}
\usepackage{graphicx}

\usepackage{amsmath}
\usepackage{amssymb}
\usepackage{pgfplots}
\usepackage{csquotes}
\usepackage{microtype}

\usepackage{color}
\usepackage[absolute]{textpos}

\usepackage[colorlinks,urlcolor=blue]{hyperref}
\usepackage[capitalize]{cleveref}

\renewcommand{\mkbegdispquote}[2]{\itshape}

\definecolor{gray92}{gray}{0.92}

\begin{document}
\title{Content-based Image Retrieval and the Semantic Gap in the Deep Learning Era}
\titlerunning{CBIR and the Semantic Gap in the Deep Learning Era}
%
\author{Björn Barz \and Joachim Denzler}
\authorrunning{B. Barz and J. Denzler}
%
\institute{Computer Vision Group, Friedrich Schiller University Jena, Jena, Germany\\
\email{\{bjoern.barz,joachim.denzler\}@uni-jena.de}}
\maketitle              
\begin{textblock*}{131.5mm}(43mm,15mm)
    \textblockcolour{gray92}
    \vspace{2mm}
    \tiny
    \centering
    This is an extended pre-print of the following article:\\
    Björn Barz and Joachim Denzler.\\
    Content-based Image Retrieval and the Semantic Gap in the Deep Learning Era.\\
    International Workshop on Content-Based Image Retrieval: where have we been, and where are we going (CBIR 2020).\\
    \copyright\ Copyright by Springer. The final publication is available at
    \href{https://link.springer.com/}{link.springer.com}.\\
    \vspace{2mm}
\end{textblock*}
\begin{abstract}
Content-based image retrieval has seen astonishing progress over the past decade, especially for the task of retrieving images of the same object that is depicted in the query image.
This scenario is called instance or object retrieval and requires matching fine-grained visual patterns between images. Semantics, however, do not play a crucial role.
This brings rise to the question: Do the recent advances in instance retrieval transfer to more generic image retrieval scenarios?

To answer this question, we first provide a brief overview of the most relevant milestones of instance retrieval.
We then apply them to a semantic image retrieval task and find that they perform inferior to much less sophisticated and more generic methods in a setting that requires image understanding.
Following this, we review existing approaches to closing this so-called semantic gap by integrating prior world knowledge.
We conclude that the key problem for the further advancement of semantic image retrieval lies in the lack of a standardized task definition and an appropriate benchmark dataset.

\end{abstract}
\section{Introduction}
\label{sec:intro}

\begin{displayquote}
    One sees well only with the heart. The essential is invisible to the eyes.
\end{displayquote}

\noindent
This famous quote from the French writer Antoine de Saint Exupéry applies to life as well as to computer vision.
The human perception of images greatly exceeds the visual surface of pixels, colors, and objects.
The \emph{meaning} of an image cannot simply be described by enumerating all objects contained therein and defining their spatial layout.
We as humans are able to grasp a plethora of diverse and complex information contained in an image at first glance, such as events happening in the depicted scene, activities performed by persons, the relationships between them, the atmosphere and mood of the image, and emotions transported by it.
Many of these concepts elude textual description and are best illustrated by providing an example image.

\begin{figure}[t]
    \includegraphics[width=\linewidth]{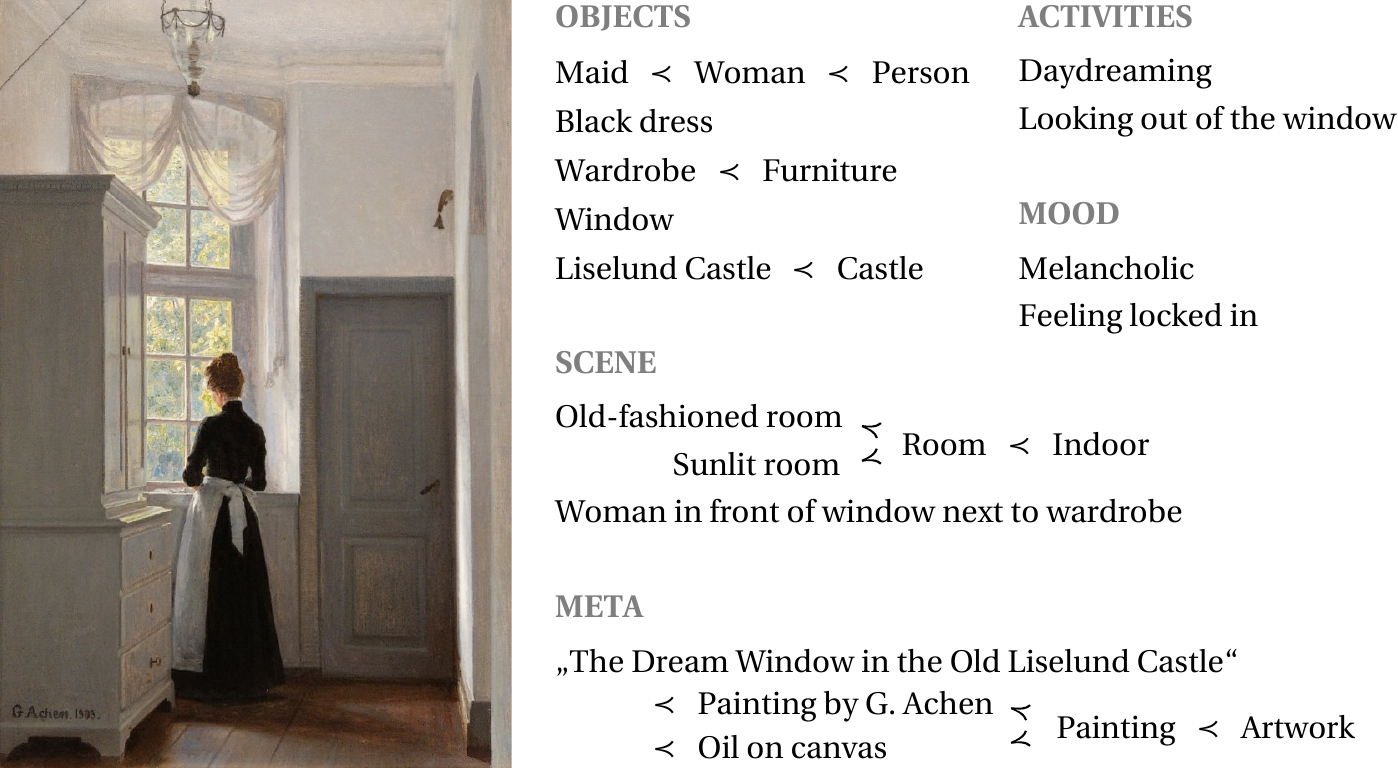}
    \caption{An example for the ambiguity and semantic richness of images. All concepts listed on the right-hand side could be used to describe the image on the left, while different observers will pay attention to different subsets of these aspects. Moreover, some concepts can be organized hierarchically, indicated by the ``$\prec$" sign, which designates the hyponomy (``is-a'') relationship.}
    \label{fig:liselund}
\end{figure}

The example in \cref{fig:liselund} illustrates this variety of information conveyed by images.
The image depicted there can be described from several perspectives: its semantic content, artistic style, the emotions it evokes in the observer, or meta-information about the image itself.
Depending on their background and the situational context, different observers will perceive and interpret this image differently.
Searching for images on the web by means of textual descriptions or keywords is hence destined to fail, because most images are not exhaustively described in their surrounding text, for mainly two reasons:
First, it is often difficult, if not impossible, to enumerate all aspects of an image explicitly, due to the potentially infinite amount of possible interpretations.
Secondly, it is not necessary to do so, since most facets of an image are directly available to the viewer by simply looking at it.
The textual description therefore focuses most often on the meta-information that is not encoded in the image itself, such as its author.
The image shown in \cref{fig:liselund}, for example, would probably be described as a photographic reproduction of the painting ``The Dream Window in the Old Liselund Castle'' by Georg Achen.
This would prevent this image from being found by users searching for images of a woman looking out of a window, images showing the activity ``daydreaming'', or images with a melancholic atmosphere.

Searching through a large database of images not with textual keywords but using a representative example as query is hence the most natural, direct, and expressive way of finding images with a particular content, which might be complex and difficult to define.
This approach is known as \emph{content-based image retrieval (CBIR)} \cite{smeulders2000cbir} and has been an active area of research since 1992 \cite{kato1992qve,niblack1993qbic}.

``Pictures have to be seen and searched as pictures'', wrote Smeulders et al.~\cite{smeulders2000cbir} in their extensive survey at the end of the ``early years'' of CBIR in 2000.
During the two decades that have passed since then, the field of content-based image retrieval has undergone at least two major revolutions (more on that in \cref{sec:instance-retrieval}).
However, most of the main challenges and directions had already been identified back then.
One of these challenges is the \emph{semantic gap}, as Smeulder et al. call it:

\begin{displaycquote}[sec.~2.4]{smeulders2000cbir}
    ``The semantic gap is the lack of coincidence between the
    information that one can extract from the visual data and the
    interpretation that the same data have for a user in a given
    situation.''
\end{displaycquote}

\noindent
Phrased with the words of de Saint Exupéry, the semantic gap is the difference between perceiving an image with the \emph{eyes}---objectively, as a depiction of objects, shapes, textures---and perceiving an image with the \emph{heart}---subjectively, including world-knowledge and emotions, reading ``between the pixels''.

The size of the semantic gap depends on the level of abstraction of the search objective pursued by the user.
Smeulders et al.~\cite{smeulders2000cbir} define this level of abstraction on a continuous scale between the two poles of a \emph{narrow} and a \emph{broad domain}.
This terminology is best explained on the basis of the three currently most relevant CBIR tasks, depicted in \cref{fig:cbir-types}:

\begin{figure}[t]
    \includegraphics[width=\linewidth]{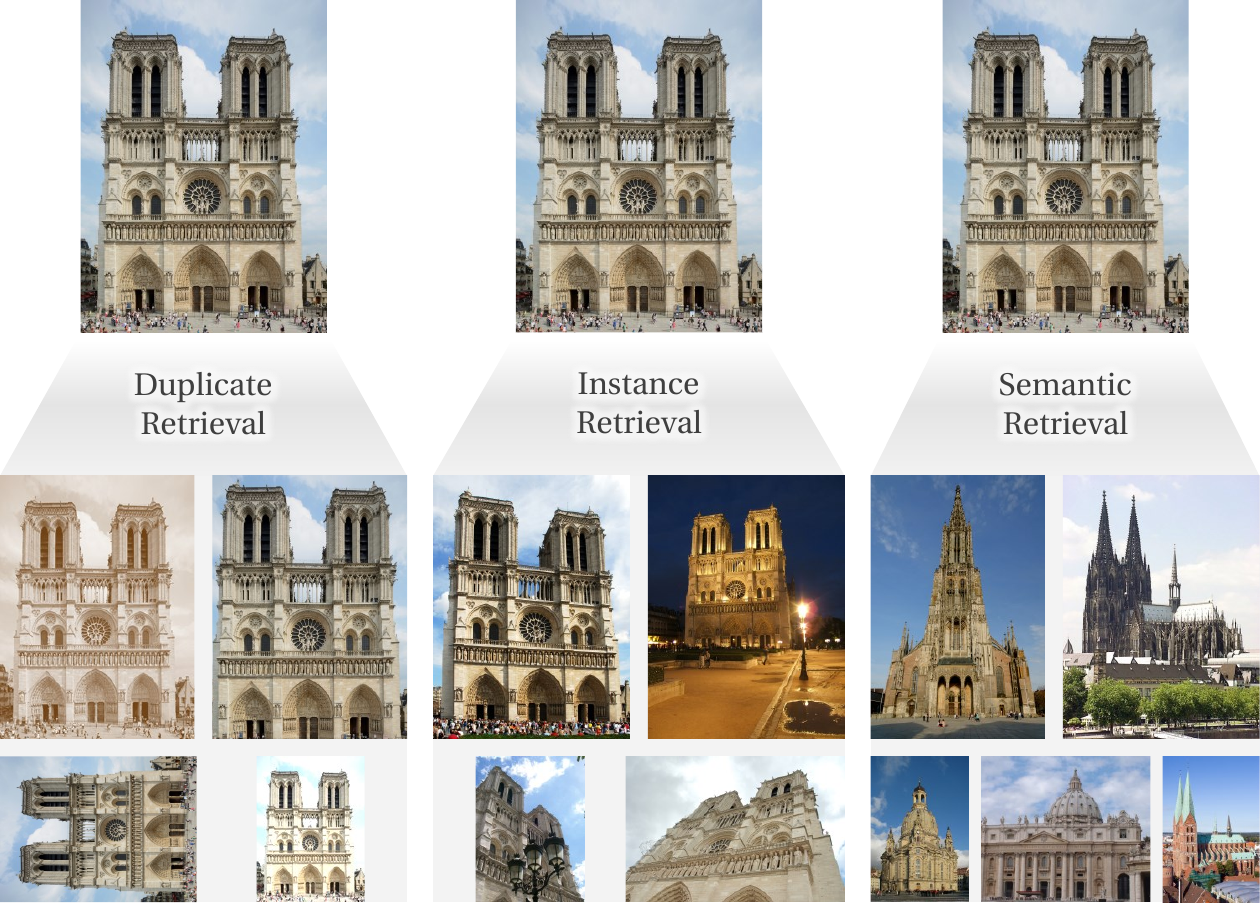}
    \caption{Examples for three different sets of images to be retrieved given the same query depending on the type of the CBIR task.}
    \label{fig:cbir-types}
\end{figure}

\begin{description}
    \item[Duplicate retrieval] searches for images with exactly the same content. These are variants that originated from the same photo but might have been post-processed differently with regard to cropping, scaling, adjustments to color, brightness, contrast etc.
    
    \item[Instance retrieval] searches for images that depict the same instance of an object, i.e., a person or a certain building. Thanks to its nature as a well-defined but non-trivial task with a clear ground-truth, this is the most extensively studied CBIR sub-task \cite{sivic2003videogoogle,perronnin2010fisher,jegou2010vlad,jegou2014triangulation,husain2017rvd,babenko2014neuralcodes,babenko2015spoc,tolias2016rmac,gordo2017deeprmac,radenovic2018finetuning,revaud2019learning,brown2020smoothap}. A handful of established datasets are available for this task \cite{jegou2008holidays,philbin2007oxford,philbin2008paris,radenovic2018oxfordrev} and significant progress has been made during the past few years, which we will outline in \cref{sec:instance-retrieval}.
    
    \item[Semantic retrieval] covers most of the remaining spectrum broader than instance retrieval and aims for finding images belonging to the same category as the query. It is important to note that \emph{category} does not necessarily mean \emph{object class} in this context. In practice, the set of possible categories is limited by nothing but the imagination of the user and a single image usually belongs to a remarkably high number of categories at once (see \cref{fig:liselund}). Thus, the exact search objective of the user can rarely be determined based on the query image alone and will almost certainly also vary between users, even for the same query. Therefore, approaches to this problem often comprise interaction with the user to adapt the similarity measure used by the system to that in the user's mind \cite{zhou2003relevance,cox2000pichunter,deselaers2008learning,Barz18:AID,Barz19:ITAL}. \\
    Learning meaningful image representations that capture fine semantic distinctions and the various facets of an image's meaning is hence of paramount importance.
    Despite its practical relevance, this CBIR sub-task has received substantially less attention than instance retrieval, mainly due to the less well-defined notion of ``relevance'' and ``similarity'' and, as a result, the lack of a suitable benchmark.
    In this work, we will review recent approaches to semantic image retrieval (see \cref{sec:semantic-retrieval}) and assess the current state of the semantic gap, twenty years after the end of the ``early years'' of CBIR.
\end{description}

\noindent
Duplicate retrieval marks one end of the spectrum, as it is the narrowest domain possible.
In this case, the semantic gap is almost non-existent and all that is needed to overcome it is a list of invariances regarding the image's content (e.g., rotation, cropping etc.).
The broader the domain, the larger the semantic gap.

While it is more challenging than duplicate retrieval, instance retrieval can still be handled by matching fine-grained distinctive visual patterns and their geometric layout.
Content-based image retrieval has made substantial progress in this area in the past two decades, which we outline in \cref{sec:instance-retrieval}.
However, the applicability of such techniques is limited with respect to the much more generic broad domain of semantic retrieval, as we see in \cref{sec:semantic-gap}.
One way to overcome this semantic gap, according to Smeulders et al.~\cite{smeulders2000cbir}, lies in integrating sources of semantic information from outside the image.
In \cref{sec:semantic-retrieval}, we review recent approaches in this direction, followed by a discussion of what is still missing for advancing CBIR in the broad domain further (\cref{sec:missing}).

\section{The Evolution of Instance Retrieval}
\label{sec:instance-retrieval}

Between 2000 and 2020, CBIR---with a particular focus on instance retrieval---has undergone two major paradigm shifts:
The first began in 2003 \cite{sivic2003videogoogle} and was initiated by the adaptation and subsequent improvement of techniques from text retrieval.
The second wave of breakthrough achievements originated from the application of deep learning methods to CBIR, starting in 2014 \cite{babenko2014neuralcodes,razavian2014cnn}.
We outline the major milestones of these two epochs of innovation in the following.

\subsection{Hand-Crafted Features and Visual Words}
\label{subsec:handcrafted}

\subsubsection{Local Features as Visual Words}

In 2003, Sivic and Zisserman \cite{sivic2003videogoogle} sought to find occurrences of a certain object in videos and, to this end, adapted the \emph{bag-of-words (BoW)} document descriptor, which is popular in the field of text retrieval, to image retrieval.
As an analogy for words, they use local image features at distinctive keypoints and quantize them into a vocabulary of ``visual words'' using the k-Means clustering algorithm.
Analogously to text retrieval, the occurrences of visual words per image are counted and the counts aggregated into a tf-idf vector representing the entire image.
Since the Euclidean distance is not meaningful in high-dimensional spaces, the cosine similarity is then used to assess the similarity of two such image representations.

This process illustrates the general framework for extracting image representations that has been used in CBIR from that point on until today \cite{jegou2014triangulation}:
A local feature extractor computes features at keypoints in a given image.
These local features are then embedded into a different space, such as quantized indices of visual words.
Finally, they are aggregated into a global representation.

The global representation allows for efficient retrieval of an initial list of candidate images.
In addition, the local features are often used to perform a spatial verification and re-ranking step for the top-ranking candidates to eliminate false matches \cite{sivic2003videogoogle,philbin2007oxford}.
This technique is quite specific to instance retrieval and matches local feature vectors between the query and a retrieved image to verify that the local features have a matching geometric layout.

\subsubsection{Towards More Complex Embeddings}

Subsequent works of this epoch focused mainly on improving the embedding and aggregation step, while using the same local feature extractor over the course of a decade.
The Hessian-affine detector \cite{mikolajczyk2004hesaff} is typically used for finding keypoints at which local features should be extracted.
This detector finds point of interest that are invariant to affine transformations as well as robust to limited changes of illumination and viewpoint.
These keypoints are then described using SIFT \cite{lowe2004sift} or RootSIFT \cite{arandjelovic2012three} features.
The latter is a simple transformation of SIFT, which consists in $L^1$-normalizing the SIFT vector and taking the element-wise square root.
In the resulting space, the Euclidean distance between RootSIFT vectors corresponds to a histogram matching kernel in SIFT space.

In the case of Sivic and Zisserman \cite{sivic2003videogoogle}, the embedding transforms each local feature vector into a space of one-hot vocabulary index vectors with tf-idf weights.
The aggregation then simply consists in a sum operation.
However, representing local feature vectors by a single integer (the cluster index), incurs a severe loss of information and does not capture the actual distribution of the local features well.
Hard assignment to a single cluster is furthermore not robust against small variations of local descriptors close to cluster boundaries.
To overcome these issues, Perronnin et al.~\cite{perronnin2010fisher} propose the use of Fisher vectors for CBIR.
The training data is quantized into visual words by fitting a Gaussian mixture model.
Each local feature vector is then transformed into the gradient of its log-likelihood with respect to the means of the Gaussians.
This realizes a weighted soft assignment to clusters and results in a dense, more informative, but also high-dimensional descriptor.
In fact, the authors show that a Fisher vector with a single visual word achieves comparable performance to a BoW descriptor with 4,000 words.

A simplification with comparable and sometimes even superior performance are \emph{vectors of locally aggregated descriptors (VLADs)}, proposed by J{\'e}gou et al.~\cite{jegou2010vlad}.
VLAD still uses hard-assignment of local descriptors to the nearest cluster, but captures the element-wise residuals of all local features from the center of their cluster.
That means, the embedding feature vector is partitioned into $k$ segments, where $k$ is the number of clusters.
The segment corresponding to the closest cluster center equals the difference between the local descriptor and that center and all other segments are 0.
The dimensionality of the embedding space is hence the number of clusters times the local feature dimensionality.
The aggregation consists in taking the sum over all transformed local feature vectors, $L^2$-normalizing the result, and applying PCA with whitening to reduce the high dimensionality of the global descriptor to something more manageable (usually in the order of a few hundred dimensions).

VLAD is, by definition, sensitive to the distance between a local feature vector and its cluster center.
However, the Euclidean distance is of limited meaning in high-dimensional spaces.
In a follow-up work, J{\'e}gou and Zisserman \cite{jegou2014triangulation} account for this fact by $L^2$-normalizing the residuals, thus encoding their angle instead of their magnitude, which gives rise to the name \emph{triangulation embedding}.
Because distance is not meaningful, hard assignments to single clusters are not reasonable either.
Triangulation embedding hence encodes the angles between the local feature vector and \emph{all} visual words.
This representation is subsequently whitened and has been found to outperform fisher vectors and VLAD.

However, Husain and Bober \cite{husain2017rvd} find that comparing each local feature vector with all visual words does not scale to large datasets.
Soft cluster assignment, on the other hand, often behaves unstable and degrades to single assignment in practice.
To overcome this, they propose a middle ground by assigning the local descriptors to the few cluster centers that are closest and base the weights on their ranks among the nearest neighbors instead of their actual distances.
These \emph{robust visual descriptors (RVDs)} are furthermore not whitened globally but on a per-cluster level.
The authors found that RVD performs competitively to triangulation embedding, while being faster to compute and more robust to dimensionality reduction.

\subsubsection{The Role of Datasets}

While the paradigm of using aggregated local features for CBIR dates back to 2003 \cite{sivic2003videogoogle}, research in this area has been most active between 2010 and 2016.
One likely reason for this delay is the lack of suitable and established benchmark datasets.
In the years 2007 and 2008, the Oxford Buildings \cite{philbin2007oxford}, Paris Buildings \cite{philbin2008paris}, and INRIA Holidays \cite{jegou2008holidays} datasets were published, which quickly emerged as the standard benchmarks for instance retrieval and gave new impetus to the field by providing a proper ground for evaluation and comparison of methods.

The two building datasets comprise different photos of various landmark buildings in Oxford and Paris, with a large variety of perspectives, scales, and occlusions.
The Holidays dataset, on the other hand, contains a collection of personal holiday photos with on average three different perspectives per scene.
While these datasets are challenging, the task of retrieving images showing the same object or scene as the query is well-defined with a clear ground truth.

\subsection{Off-the-shelf CNN Features}
\label{subsec:cnn-feat}

After hand-crafted local features had remained unquestioned in CBIR for over a decade, the renaissance of deep learning finally led to a substantial change regarding image representations.
The independent works of Babenko et al.~\cite{babenko2014neuralcodes} and Razavian et al.~\cite{razavian2014cnn} first showed that surprisingly good results can be achieved by simply extracting global image descriptors, so-called \emph{neural codes}, from the first fully-connected layer of an off-the-shelf CNN pre-trained on ImageNet \cite{deng2009imagenet}.
Given the extreme simplicity of this approach, requiring close to zero engineering effort compared to detecting keypoints, extracting local features, and aggregating them, this was a remarkable result.
Just a year later, Babenko and Lempitsky~\cite{babenko2015spoc} considerably improved the performance of this approach by extracting image features not from a fully-connected but from the last convolutional layer, which still has a spatial resolution.
The result is, thus, a set of feature vectors, which can roughly be associated with different regions in the image.
These are summed up for aggregation, $L^2$-normalized, reduced in dimensionality using PCA, and $L^2$-normalized again, leading to the speaking name \emph{sum-pooled convolutional features (SPoC)} for these descriptors.

In the following years, research mainly adhered to using such pre-trained neural feature extractors and focused on designing sophisticated aggregation functions.
Many of them try to find a middle ground between sum and maximum pooling, e.g., by averaging activations over the top few responses only as in \emph{partial mean pooling (PMP)} \cite{zhi2016pmp}, or by smoothly interpolating between the two extremes as in \emph{generalized-mean pooling (GeM)} \cite{radenovic2018finetuning}.

Aggregated convolutional features have one drawback, though: As opposed to traditional local features, they do not allow for precise localization of the matching object and, thus, are not compatible with techniques such as spatial verification and re-ranking, which depend on geometric information.
To this end, Tolias et al.~\cite{tolias2016rmac} propose the \emph{regional maximum activation of convolutions (R-MAC)} aggregation, which follows a two-step approach:
The convolutional feature map is divided into overlapping regions of different sizes and the local feature vectors in each region are aggregated using maximum pooling.
These so-called MAC vectors are then whitened and aggregated by sum pooling into a global R-MAC image descriptor.
For spatial re-ranking, the similarity of the query's MAC vector and the individual regional MAC vectors of the top few retrieval results can be used to localize the query object in the retrieved images and refine the ranking.

These techniques took CBIR based on features extracted from pre-trained CNNs quite far, but the hand-crafted RVD descriptor \cite{husain2017rvd} is still able to compete with them on instance retrieval benchmarks.

\subsection{End-to-end Learning for Image Retrieval}
\label{subsec:end2end}

Deep learning finally became undeniably superior to traditional CBIR techniques based on hand-crafted features when researchers began to adapt the CNN used for feature extraction to the task of image retrieval instead of using a pre-trained one.
We regard this shift of focus from feature transformation and aggregation to actual feature learning as the second important paradigm shift in CBIR.

\subsubsection{Global Features}

Gordo et al.~\cite{gordo2017deeprmac} were among the first to be successful in this endeavor and set the state of the art in instance retrieval for at least two years.
They build upon R-MAC \cite{tolias2016rmac} and implement it as differentiable layers on top of a VGG16 CNN architecture, which can then be trained end-to-end.
To this end, they employ the triplet loss \cite{schroff2015facenet}, a training objective from the field of deep metric learning.
By training on a curated dataset of famous landmarks, they learn a feature representation where images of the same landmark are closer together by a certain margin than two images of different landmarks, which supports the objective of instance retrieval.

This approach has later been extended by extracting R-MAC features from multiple layers of a CNN and weighting individual features of each region by the Kullback-Leibler divergence between the distributions of the Euclidean distance between matching and non-matching descriptors, so that more discriminative regional features obtain a higher weight \cite{husain2019remap}.
The motivation for combining features from multiple layers lies in the different degrees of visual abstraction: features from earlier layers are more indicative of visual properties, while later layers provide a semantically more abstract representation.

As opposed to the triplet loss, Radenovi{\'c} et al.~\cite{radenovic2018finetuning} find the contrastive loss to provide better final performance, while furthermore requiring only pairs instead of triplets of images for training.
More importantly, they propose an unsupervised technique for generating training data consisting of matching and non-matching image pairs for instance retrieval without human annotation:
Images in the training dataset are clustered based on their BoW representation using local RootSIFT features and spatial verification is applied to ensure that all images in a cluster show the same object.
A 3-D model is then constructed for each cluster using structure-from-motion (SfM) techniques, so that it can be determined from these models whether two images depict the same object or not.
This also allows images of the same landmark but captured from different and disjoint viewpoints to be considered as non-matching.
The information about camera positions obtained from SfM furthermore enables mining of challenging positive image pairs that exhibit a non-trivial amount of overlap.

These metric learning approaches have led to an impressive improvement of instance retrieval performance in terms of average precision (AP), even though they do not optimize it directly but a proxy objective based on distances in the learned feature space.
Since AP is the most important metric for evaluating retrieval methods, it seems desirable to optimize it directly instead of a proxy-task.
However, that entails taking into account not only a single sample, a pair, or a triplet as before, but the entire list of ranked results.
One apparent benefit is that such listwise objectives are position-sensitive: The impact of a single pair or triplet involving images at the top of the ranking should be higher than at the end of the list.
However, average precision is not differentiable, because it involves sorting images by their similarity to the query.
For being able to optimize AP in an end-to-end learning context nevertheless, He et al.~\cite{he2018local} proposed a differentiable approximation of AP using histogram binning, which has been adopted by Revaud et al.~\cite{revaud2019learning} for CBIR and improved the state of the art.
Since the cosine similarity, which is usually employed for retrieval, is bounded in $[-1,1]$, the range of possible similarity scores can easily be divided into a fixed number of equally sized bins.
Images are then soft-assigned to the bins whose centers are closest to the image's retrieval score to obtain histograms of positive and negative match counts in each bin.
Instead of computing precision and recall for each possible position in the ranking, these metrics can now be computed for each bin and combined to approximate AP.

However, the quantization of similarity scores into bins ignores variations of the ranking within each bin, which can have particularly large impacts on AP at the top positions of the ranking.
This deficiency has recently been overcome by a different approach to approximating AP:
Instead of quantized sorting by binning, the sorting operation itself is relaxed by replacing the Heaviside step function indicating whether one element of the list precedes another with a sigmoid function to avoid vanishing gradients \cite{prillo2020softsort}.
This allows for differentiable sorting and computation of a relaxed version of AP, called Smooth-AP \cite{brown2020smoothap}.

With these listwise approaches, global representations for CBIR can finally be learned end-to-end without hand-crafted intermediate steps or proxy objectives.

\subsubsection{Local Features}

While global image descriptors are convenient for retrieval applications, they are neither robust in the presence of occlusion or background clutter nor suitable for spatial verification, which is an important technique for instance retrieval.
Other works hence aimed at learning local feature detectors and descriptors in an end-to-end manner.

\emph{Deep Local Features (DELF)} \cite{noh2017delf}, for example, uses coarse regional features extracted from a convolutional layer of a pre-trained CNN and then trains another small CNN to assess the importance of these densely sampled keypoints.
For training, these predicted weights are used for weighted sum pooling of the local descriptors into a global feature vector, which allows for fine-tuning of the local features using image-level supervision.

Most instance retrieval systems using local features adopt a two-stage approach:
First, a set of candidate images is retrieved by comparing global features and then re-ranked using spatial verification based on local features.
Cao et al.~\cite{cao2020delg} unified the learning of both types of features into a single model with two branches:
One branch aggregates all feature vectors of the last convolutional layer of a CNN as global feature vectors and is trained with a metric learning loss.
The other branch learns an attention module to identify distinctive local features and is trained using categorical cross-entropy.

\subsubsection{The Need for More Challenging Benchmarks}

Besides plenty of computing capacity, deep learning techniques require one thing most of all: data.
The existing instance retrieval datasets were too small for training deep neural networks, wherefore Babenko et al.~\cite{babenko2014neuralcodes} created a novel landmarks dataset with over 200,000 images for training purposes, which was later used by other works on deep image retrieval as well \cite{gordo2017deeprmac}.
Nowadays, the large-scale Google-Landmarks dataset \cite{noh2017delf} proposed in 2017 is often used for training.
It comprises over a million images of 12,894 landmarks from all over the world.

These datasets are orders of magnitudes larger than the Oxford and Paris Buildings dataset, but the latter were still relevant for evaluating and comparing novel methods.
The rapid advances in deep learning for CBIR, however, quickly resulted in a saturation of performance on these benchmarks.
Therefore, Radenovi{\'c} et al.~\cite{radenovic2018oxfordrev} revisited these two datasets in 2018 by improving the ground-truth annotations, finding more difficult queries, adding challenging distractor images, and defining three different evaluation protocols of varying difficulty.

These developments demonstrate the importance of suitable training and benchmark datasets for the advancement of content-based image retrieval.

\section{Impact on the Semantic Gap}
\label{sec:semantic-gap}

The previous section outlined the impressive advances of instance retrieval in the deep learning era.
However, instance retrieval is a rather narrow domain, where a broad understanding of the scene semantics are not required to solve the task satisfactorily.
The interesting question is, therefore: Do these advances transfer to the broader domain of semantic retrieval?

To answer this question, we evaluate several seminal methods and models on an instance retrieval and a semantic retrieval task.
For instance retrieval, we use the Revisited Oxford Buildings dataset \cite{philbin2007oxford,radenovic2018oxfordrev} (see above), on which these methods have originally been evaluated.
As an indicator for their performance in a broader domain, we evaluate them on the MIRFLICKR-25K dataset \cite{huiskes2008mirflickr}, which comprises 25,000 images from Flickr, each annotated with a subset of 25 concepts such as ``sky'', ``lake'', ``sunset'', ``woman'', ``portrait'' etc.
While most images in the dataset are annotated with more than one concept, 3,054 of them exhibit only a single label.
We use these images as queries to avoid query ambiguity.
We consider a retrieved image as relevant if it shares this concept.

\begin{figure}[t]
    \resizebox{\linewidth}{!}{\input{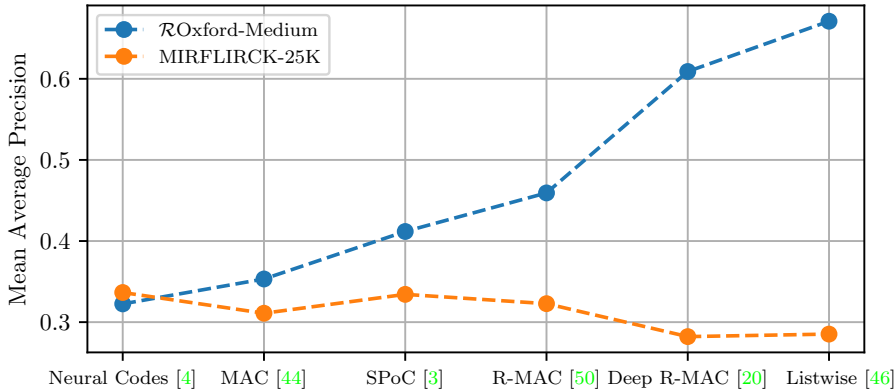}}%
    \caption{Milestones of CNN-based instance retrieval, evaluated on an instance retrieval ($\mathcal{R}$Oxford \cite{radenovic2018oxfordrev}) and a semantic retrieval dataset (MIRFLICKR-25K \cite{huiskes2008mirflickr}).}
    \label{fig:performance-evolution}
\end{figure}

\Cref{fig:performance-evolution} depicts the mean average precision of several milestones of CBIR research in the deep learning era on both tasks.
While the performance on instance retrieval tasks increased steadily, the semantic retrieval performance did not only not improve, but even deteriorated slightly.
The majority of developments in the past years have focused on instance retrieval and hence tuned feature representations towards this tasks, for which fine-grained visual features are important.
This, however, degraded their performance on broader-domain tasks, for which a different set of features is necessary.

While instance retrieval has reached a very advanced level of maturity during the past 20 years, content-based image retrieval in general is still facing the challenges of the semantic gap.

\section{Knowledge Integration for Semantic Image Retrieval}
\label{sec:semantic-retrieval}

One way to overcome the semantic gap lies in incorporating additional sources of information outside the image, as Smeulders et al.~\cite{smeulders2000cbir} already stated back in 2000.
In the following, we briefly review the most common sources of such external information as well as approaches for leveraging them to improve image representations for semantic image retrieval.

\subsection{Class Labels}
\label{subsec:class-labels}
%

Image-level class labels are one of the most frequently available and cheapest types of semantic information about images.
To provide robust performance in an open world, however, a huge number of classes or sophisticated methodology beyond training a simple classifier is required.

\emph{OASIS} \cite{chechik2010oasis} combines both:
Method-wise, OASIS learns a bilinear similarity metric using the triplet loss for comparing hand-crafted features with respect to semantic image similarity.
The training dataset consisted of over two million images sourced from Google Image Search using about 150,000 textual queries entered by real users.
Working at Google, the authors did not only have access to these queries, but also to relevance ratings based on click statistics, which allowed them to collect this large-scale but non-public dataset.

With the advent of deep learning, Yu et al.~\cite{yu2017exploiting} exploit the intrinsic hierarchical representation generated by CNNs by combining features from shallow and deep layers.
While the former capture rather visual patterns, features from deeper layers are expected to be more abstract and carry semantic information.
Despite this, they only evaluate their approach on instance retrieval benchmarks.

More recent approaches optimize CNNs directly for multiple tasks to learn diverse representations.
\emph{MultiGrain} \cite{berman2019multigrain}, for instance, aims for learning features that are useful for class-level, instance-level, and identity-level recognition by combining a classification and a metric learning objective.
Evaluation, however, is conducted separately for each task in terms of classification accuracy on ImageNet \cite{deng2009imagenet} and retrieval accuracy on instance retrieval benchmarks.
This evaluation protocol does not provide information about semantic retrieval performance.

To deploy CBIR at production-level within the \emph{Microsoft Bing} search engine, Hu et al.~\cite{hu2018bing} employ a large ensemble of different network architectures trained for various tasks: for classification with cross-entropy loss, with a metric learning objective such as the contrastive or triplet loss, for face recogmition, or for object detection.
This ensemble is intended to capture a broad variety of both visual and semantic properties of images and, hence, cover most objectives a user of the visual search engine could pursue.
The training data for this system is non-public and was collected by human annotators in an expensive data collection and annotation effort.
The evaluation was conducted using human relevance judgments as well.
For these two reasons, this work is neither publicly reproducible nor directly comparable with other works.

\subsection{Class Taxonomies}
\label{subsec:taxonomies}
%

Plain class labels do not take into account the semantic relationships between classes.
Despite their visual similarity, images of humans and apes, for example, are generally considered to be semantically much less similar than images of a caterpillar and a butterfly, although the latter are not particularly similar from a visual perspective.
Taxonomies such as WordNet \cite{fellbaum1998wordnet} are a popular tool for measuring the semantic similarity between classes.
They organize concepts on different levels of abstraction in terms of is-a relationships (``a poodle is a dog is an animal etc.'').
Several works strive for integrating this prior knowledge about the world to improve the semantic consistency of CBIR results.

Deng et al.~\cite{deng2011hierarchical} construct a hand-crafted bilinear similarity measure from the class taxonomy of ImageNet \cite{deng2009imagenet} and use it for comparing vectors of class probabilities predicted by a classifier.
Instead of a similarity measure, Barz and Denzler~\cite{Barz19:Embeddings} construct a semantic feature space spanned by class embeddings, where the cosine similarity between two class embeddings equals their semantic similarity derived from the taxonomy.
They then use a CNN to map images into the same semantic space.
Arponen and Bishop~\cite{arponen2019shrewd} do not constrain the feature space in this explicit way, but instead integrate the same objective directly into the loss function, so that the layout of the semantic feature space is learned.
They combine this with an additional term encouraging the individual features to be binary, which allows for compact and memory-efficient descriptors.

The aforementioned works evaluate their approaches on ImageNet using ``hierarchical precision'' \cite{deng2011hierarchical}, which replaces the binary relevance of the retrieval results used by ordinary precision with the semantic similarity of their class and the class of the query.
This metric suits the task better, but is best plotted for several cut-off positions in the ranking and cannot easily be summarized in a single number to facilitate comparison.

Yang et al.~\cite{yang2019adaptive} combine semantic and visual similarity by first ranking images according to semantic similarity and then ordering the images within the same class according to visual similarity to the query.
To this end, they use the contrastive loss with an adaptive margin proportional to the dissimilarity.
The evaluation, however, is limited to fine-grained classification datasets and conducted using binary relevance, which does not take semantics into account.

Long et al.~\cite{long2020searching} not only embed the classes but all concepts in the taxonomy into a hyperbolic space, so that sub-classes lie in their parent class' entailment cone.
As before, a CNN is then used to map samples onto their class embeddings.
Although their method could also be applied for content-based image retrieval, they focus on video retrieval and evaluate their approach on that task only.

\subsection{Textual Descriptions}
\label{subsec:text}
%

While taxonomies provide information about the semantic similarity between classes, their full semantic meaning goes far beyond that.
Several works have aimed for extracting such rich semantics from textual descriptions of classes or images and leverage them for learning meaningful image features.
\emph{DeViSE}~\cite{frome2013devise} and \emph{HUSE}~\cite{narayana2019huse}, for example, learn word embeddings on Wikipedia and use the embedding of a class' name as its semantic embedding.
DeViSE~\cite{frome2013devise} then maps images into that space by maximizing the dot-product similarity between their feature vector and the respective class embedding, while enforcing a certain minimum distance to any other class embedding.
HUSE~\cite{narayana2019huse}, in contrast, adopts a pair-wise optimization approach by forcing the distance of pairs of images to be equal to the dissimilarity of their class embeddings.
This approach provides more flexibility regarding the learned image feature space since it is separate from the space of word embeddings.
Like some of the hierarchy-based approaches described above, both methods were evaluated using hierarchical precision.
Thus, the semantic information used for evaluation was not the same as that used for training, which incurs a disadvantage compared to hierarchy-based methods.

Instead of using texts associated with classes, other methods leverage texts belonging to individual images, such as titles and captions, and learn a multi-modal embedding space.
Gomez et al.~\cite{gomez2018learning} do so by training a CNN to regress the text embeddings generated by a separately trained language model.
However, they evaluate their approach only with textual queries and not in a \emph{content-based} image retrieval scenario.
Wu et al.~\cite{wu2019uvse}, in contrast, learn text and image embeddings jointly and additionally predict individual embeddings for components of the caption such as objects, object-attribute pairs, and object-relation phrases.
These semantic components are automatically aligned with the local features of the corresponding image regions using contrastive learning.
However, their experiments only investigate the cross-modal image-to-caption and caption-to-image retrieval scenarios, while semantic CBIR performance is not analyzed.

\subsection{Artistic Style}
\label{subsec:style}
%

An entirely different dimension of image semantics is opened up by stylistic concepts such as artistic style, mood, and atmosphere.
Learning image features that respect such properties requires either specialized annotations or prior knowledge about their characteristics.

Ha et al.~\cite{ha2020color} define style in terms of color composition, i.e., the distribution and layout of colors in an image.
They construct a dataset with subjective 5-star similarity ratings for pairs of images, which have been collected in a laborious crowd-sourcing process involving active learning.
A siamese network is trained to predict the distribution of similarity ratings for a given pair of images.

To avoid the expensive collection of large-scale style datasets, Gairola et al.~\cite{gairola2020style} draw on knowledge from the field of visual style transfer, where Gram matrix features have been found to capture the stylistic properties of images.
They extract these features from a pre-trained CNN, cluster them, and use the cluster labels as ground-truth for training another CNN using the triplet loss.
They evaluate their approach on numerous datasets annotated with artistic styles, photographic styles, historical art styles, moods, or genres.

\section{The Missing Ingredient}
\label{sec:missing}
%

The two lines of research on instance retrieval and semantic retrieval portrayed in \cref{sec:instance-retrieval,sec:semantic-retrieval}, respectively, exhibit one apparent difference:
The research on instance retrieval shows measurable continuous progress thanks to the Oxford~\cite{philbin2007oxford} and Paris~\cite{philbin2008paris} benchmark datasets, whose release was followed by a clear surge of research activity in the field.
With the Google-Landmarks dataset~\cite{noh2017delf}, sufficient training data is available for modern deep learning methods.
The recent revision of the two aforementioned benchmark datasets \cite{radenovic2018oxfordrev} maintains their usefulness as a benchmark despite the substantial performance improvements.

Existing works on semantic image retrieval, in contrast, vary widely with respect to their evaluation protocol, training data (some of which is closed-source), and even the problem definition, rendering a clear comparison between approaches impossible.
This is perhaps the biggest obstacle for further progress in this field and the likely reason why research still focuses on instance retrieval.

A thoroughly curated benchmark dataset for semantic image retrieval would hence greatly contribute to advancing the field.
However, constructing such a benchmark is highly non-trivial due to numerous aspects.
This begins already with the evaluation metric.
In a semantic CBIR scenario, precision is often more important than recall, since most users are not interested in all potentially relevant images from a large-scale database.
Average precision is hence a sub-optimal measure, but also precision alone is insufficient, since it only considers binary relevance.
In reality, however, relevance is a graded phenomenon \cite{smeulders2000cbir}.
A candidate for an evaluation metric is the normalized discounted cumulative gain (NDCG) \cite{jarvelin2002cumulated}, which is capable of taking into account the degree of relevance between two images.
The dataset, however, also needs to provide such graded relevance ratings for each pair of query and retrieved image.
Ideally, the relevance should be based on real user ratings, which poses a major annotation effort.

Furthermore, the benchmark should define a diverse set of relevance criteria a user can have in mind when using a CBIR system, including instance identity, object category identity on different levels of abstraction, similarity regarding artistic style, mood, emotions, actions, and relationships portrayed in the image.
Further complications are caused by the fact that a single query image can be interpreted differently with respect to each of these dimensions.
The relevance of a retrieved image hence does not only depend on the query, but also on the search objective pursued by the user.
This ambiguity can only be resolved by interaction with the user or by providing multiple query images sharing the relevant aspect.
Therefore, the benchmark should ideally provide different evaluation protocols, an interactive one and a non-interactive one, which could be restricted to less ambiguous queries.
The interactive scenario furthermore requires the definition of a feedback simulation protocol.

\section{Conclusions}
\label{sec:conclusions}

Content-based image retrieval has made astounding progress over the past two decades, especially in the area of instance retrieval, where a clearly defined objective and evaluation benchmarks exist.
However, the methodological advances in this area do not translate to the more challenging task of semantic image retrieval.
On the contrary, more advanced instance retrieval methods often perform worse than simpler ones in that domain.
Despite the seeming advances, the semantic gap has rather become larger than smaller.

Due to the lack of an established benchmark, semantic image retrieval methods are often hardly comparable and vary widely regarding the task definition and the evaluation data and protocol.
The history of instance retrieval shows that such a benchmark would be an invaluable catalyst for research on semantic image retrieval and a necessity for closing the semantic gap.

%
%
\bibliographystyle{splncs04}
\bibliography{references}

\end{document}